Forecasting COVID-19 Infections in Gulf Cooperation Council (GCC) Countries using Machine Learning

Leila Ismail*

Intelligent Distributed Computing and Systems Research Laboratory, Department of Computer Science and Software Engineering, College of Information Technology, United Arab Emirates University, Al Ain, Abu Dhabi, 15551, United Arab Emirates

Huned Materwala

Intelligent Distributed Computing and Systems Research Laboratory, Department of Computer Science and Software Engineering, College of Information Technology, United Arab Emirates University, Al Ain, Abu Dhabi, 15551, United Arab Emirates

Alain Hennebelle

Independent Data Scientist and Engineer, Al Ain, Abu Dhabi, United Arab Emirates

COVID-19 has infected more than 68 million people worldwide since it was first detected about a year ago. Machine learning time series models have been implemented to forecast COVID-19 infections. In this paper, we develop time series models for the Gulf Cooperation Council (GCC) countries using the public COVID-19 dataset from Johns Hopkins. The dataset set includes the one-year cumulative COVID-19 cases between 22/01/2020 to 22/01/2021. We developed different models for the countries under study based on the spatial distribution of the infection data. Our experimental results show that the developed models can forecast COVID-19 infections with high precision.

# 1  Introduction

The novel coronavirus (COVID-19) was declared as a global pandemic by the World Health Organization (WHO) after it was first discovered in Wuhan, China [1]. Over one year, the virus has infected more than 68 million people worldwide [2]. The virus can be fatal for elderly people or ones with chronic diseases [3]. Different countries across the globe have imposed several social practices and strategies to reduce the spread of the infection and to ensure the well-being of the residents. These practices and strategies include but are not limited to social distancing, restricted and authorized travels, remote work and education, reduced working staff in organizations, and frequent COVID-19 tests. These measures have been proved potential in reducing the disease spread and death in the previous pandemics [3], [4].

Several studies have focused on machine learning time series models to forecast the number of COVID-19 infections in different countries [5, 6, 7, 8, 9, 10, 11, 12, 13, 14]. This is to aid the government in designing and regulating efficient virus spread-mitigating strategies and to enable healthcare organizations for effective planning of health personnel and facilities resources. Based on the forecasted infections, the government can either make the confinement laws stricter or can ease them. Similarly, the hospitals can provision more beds, ventilators and/or medications if the time series model forecasts a rise in infections. The machine learning time series models are developed based on past COVID-19 infections to forecast future infections.

*Correspondence: Leila Ismail (email: leila@uaeu.ac.ae)

In this paper, we develop machine learning time series models to forecast the COVID-19 infections in the Gulf Cooperation Council (GCC) countries. GCC countries include the United Arab Emirates (UAE), Saudi Arabia, Qatar, Oman, Kuwait, and Bahrain. We develop the time series models for these countries using the infections dataset collected from the Johns Hopkins COVID-19 dataset [15]. To obtain reliable infections forecasts, we develop a time series model for each country under study based on the spatial distribution of infections data for that country. This distribution may differ from one country to another depending on the geographical characteristics, social behaviors, confinement strategies and the residents' response towards the strategies being implemented. Therefore, the use of a single model to forecast infections growth for all the countries leads to inaccurate results [16]. This is because the model that embraces the data trend would accurately fit the data distribution under modeling and give better precision that time series models. We develop the Autoregressive Integrated Moving Average (ARIMA) model [17] for UAE and Bahrain, the Holt's Linear Trend (HLT) model [18] for Qatar, Oman and Kuwait, and the Damped Trend (DT) model [19] for Saudi Arabia. The developed models will aid the government and healthcare organizations to estimate infection growth. We evaluate the performance of the developed models in terms of root mean squared error (RMSE) and mean absolute percentage error (MAPE). RMSE does not allow the performance comparison of the developed models between countries with heterogeneous scale of infections' spread. Consequently, MAPE is also considered for evaluation.

The main contribution of this paper is to model COVID-19 infections spread for the GCC countries using different time series models based on the infection data distribution. The models are developed using cross-sectional study data of 367 observations for six different countries. The result of this paper will guide the policymakers towards the development of more effective pandemic precautionary measures.

The rest of the paper is organized as follows. Section 2 overviews the related work. The dataset used for the models' development, the analysis of the infections data distribution trend, and the time series models are presented in Section 3. Section 4 discusses the experiments and the analysis of the results. In Section 5 we conclude the paper.

## 2 RELATED WORK

Several studies have focused on machine learning time series models to forecast the number of COVID-19 infections in different countries including the ones in GCC [12, 13, 14]. Hernandez-Matamoros et al. [12] implemented the ARIMA model to forecast COVID-19 infections for 175 countries. They justified the selection of ARIMA by its accuracy when applied to other countries in the literature. In this paper, we select for each GCC country the model which is more aligned with the trend of the infection's growth [16]. Similarly, Abuhasel et al. [14] selected ARIMA to forecast the infections in Saudi Arabia, and Al-Sharoot and Alwan [13] selected ARIMA for UAE and Saudi Arabia, based on the literature. In this paper, we use ARIMA for UAE and DT for Saudi Arabia based on their respective data trends. The developed models in this paper will aid the government and healthcare organizations to estimate infection growth. This will aid the government to effectively plan confinement policies and strategies and will aid healthcare organizations to efficiently organize personnel and medical resources.

## 3 METHODOLOGY

### 3.1 Dataset

The dataset used in this paper is created from the publicly available Johns Hopkins COVID-19 dataset. We extracted the number of confirmed cases data for the considered countries, i.e., UAE, Saudi Arabia, Qatar, Oman, Kuwait, and Bahrain. The extracted data includes the one-year cumulative COVID-19 cases between 22/01/2020 to 22/01/2021. Figure 1 shows the infection trend for GCC countries. It shows that the infection distribution for the UAE and Bahrain has an exponential trend over the period considered. The infection distribution for Qatar,

Oman, and Kuwait is a combination of exponential and linear trends, and the infection distribution for Saudi Arabia depicts an exponential trend with damping. A study by [16] reveals that a single time series model cannot be implemented to forecast COVID-19 infections for countries exhibiting different infection distributions. Based on these results, the ARIMA time series model is best suited for infection forecasting if the data distribution is exponential, the HLT model is the best suited if the data distribution is exponential + linear and the DT model is the best suited if the infections data distribution is exponential damping.

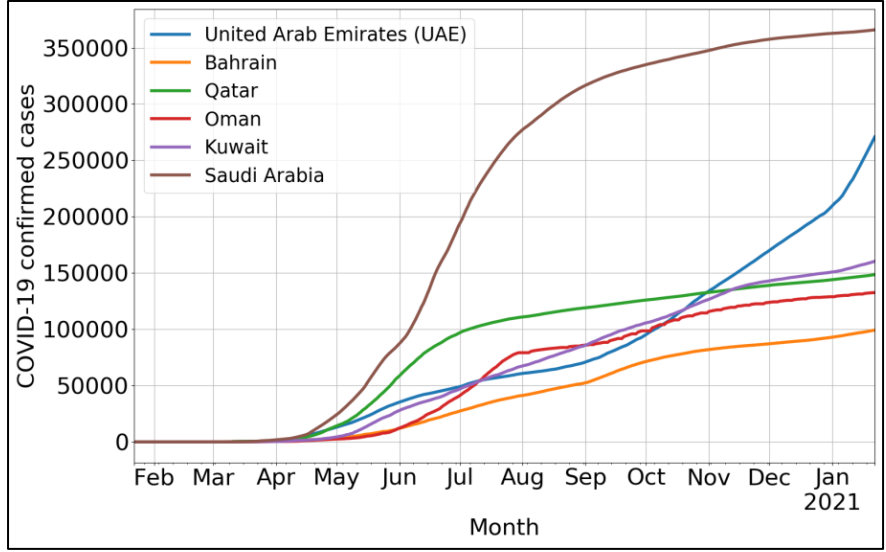

Figure 1: COVID-19 infections data trend for GCC countries

## 3.2 Time Series Models

### 3.2.1 Autoregressive Integrated Moving Average.

Autoregressive integrated moving average (ARIMA) model is an amalgam of the autoregressive (AR) and the moving average (MA) models [17]. AR is a time series method that develops a linear regression model to forecast the COVID-19 infections as a function of the lagged (previous) infections as stated in Equation (1). MA method, also known as the rolling mean, develops a linear model to forecast the COVID-19 infections based on the lagged (previous) forecast errors as stated in Equation (2). ARIMA method can only be applied to a stationary data trend. Therefore, a non-stationary data trend is transformed into a stationary one by the process of differencing. In differencing, a data value is replaced by the difference between that value and the previous value. In the ARIMA model, the time series is first differenced to make it stationary and then the AR and MA methods are applied. Consequently, the differenced infections data trend is combined with the AR and MA methods as stated in Equation (3).

$$Infections_T = \alpha + \beta_1 Infections_{T-1} + \beta_2 Infections_{T-2} + \cdots + \beta_p Infections_{T-p} + \varepsilon_T \quad (1)$$

$$Infections_T = \alpha + \varepsilon_T + \gamma_1 \varepsilon_{T-1} + \gamma_2 \varepsilon_{T-2} + \cdots + \gamma_q \varepsilon_{T-q} \quad (2)$$

$$Infections'_T = \alpha + \beta_1 Infections'_{T-1} + \beta_2 Infections'_{T-2} + \cdots + \beta_p Infections'_{T-p} + \gamma_1 \varepsilon_{T-1} + \gamma_2 \varepsilon_{T-2} + \cdots + \gamma_q \varepsilon_{T-q} + \varepsilon_T \quad (3)$$

where α, β, and γ are the regression coefficients, $\varepsilon_T$ represents the infections forecast error at a given time T, and Infections'$_T$ is the number of infections at a given time T in the differenced series. The series may have been differenced more than once to obtain a stationary time series.

The lag period for the AR method (p), the lag period for the MA method (q), and the order of the differencing (d) are considered as the ARIMA parameters. These parameters' values are required to be determined before implementing the ARIMA method for forecasting COVID-19 infections.

### 3.2.2 Holt's Linear Trend.

Holt's Linear Trend (HLT) extends the simple exponential smoothing time series model that enables the forecasting of infections with a trend [18]. The infections forecast model for HLT is a linear function of the forecast period as stated in Equation (4).

$$Infections_{T+h|T} = l_T + hb_T \qquad (4)$$

where $l_T$ represents the smoothing of the level in the series at a given time T, $b_T$ represents the smoothing of the trend in the series at a given time T and h represents the forecast period. The values of $l_T$ and $b_T$ can be computed as shown in Equations (5) and (6) respectively.

$$l_T = (\alpha)(Infections_T) + (1-\alpha)(l_{T-1} + b_{T-1}) \qquad (5)$$
$$b_T = (\beta^*)(l_T - l_{T-1}) + (1-\beta^*)(b_{T-1}) \qquad (6)$$

where α and β* are the smoothing parameters for level and trend, respectively such that α, β* ϵ [0,1] and Infections T represents the number of COVID-19 infections at time T.

### 3.2.2 Damped Trend.

The HLT model discussed previously either forecasts an indefinitely increasing or decreasing infections trend. This may lead to an over-forecast for a country where the increment in the number of COVID-19 infections dampens after some time in the future. The DT time series model addresses this issue by including a damping parameter in the HLT model [19]. This dampens the steep HLT forecast to a flat trend at some point of time in the future. The infections forecast model for the DT model is stated in Equation (7).

$$Infections_{T+h|T} = l_T + (\emptyset + \emptyset^2 + \cdots + \emptyset^h)b_T \qquad (7)$$

where Ø is the damping parameter such that Ø ϵ (0,1). The DT method behaves similarly to the HLT method when Ø = 1. $l_T$ and $b_T$ are computed using Equations (8) and (9) respectively.

$$l_T = (\alpha)(Infections_T) + (1-\alpha)(l_{T-1} + \emptyset b_{T-1}) \qquad (8)$$
$$b_T = (\beta^*)(l_T - l_{T-1}) + (1-\beta^*)(\emptyset b_{T-1}) b_T = (\beta^*)(l_T - l_{T-1}) + (1-\beta^*)(b_{T-1}) \qquad (9)$$

where α and β* are the smoothing parameters for level and trend, respectively such that α, β* ϵ [0,1].

## 4 PERFORMANCE ANALYSIS

In this section, we discuss the experiments performed to develop the models under study and present the analysis of the experimental results. We evaluate the performance of the models in terms of RMSE and MAPE.

## 4.1 Experiments

For each considered country, we use the confirmed cases features of the Johns Hopkins data to create a dataset for that country. For each country, we use 70% of the dataset 22 January – 03 October 2020) for training, i.e., to develop the model corresponding to that country, and 30% of the dataset (04 October – 22 January 2021) for validation, i.e., to evaluate the performance of the developed model.

To validate a developed model, we forecast the COVID-19 infections for the corresponding country during 04 October – 22 January 2021 and then compare the forecasted infections with the actual number of infections. We evaluate the performance of the models in terms of RMSE and MAPE. The RMSE and MAPE values are computed using Equations (10) and (11) respectively.

$$RMSE = \sqrt{\frac{\sum_{T=1}^{h}(Infections_T - Infections'_T)^2}{h}} \qquad (10)$$

$$MAPE = \left(\frac{1}{h}\sum_{T=1}^{h}\frac{|Infections_T - Infections'_T|}{Infections_T}\right) * 100\% \qquad (11)$$

where $Infections_T$ and $Infections'_T$ are the actual and forecasted infections for the duration of h period.

To obtain the values of α and β parameters for the HLT and DT models, we run the models with varying values of the parameters between 0 and 1 at an interval of 0.1, i.e., (α =0, β =0), (α =0, β =0.1), …, (α =0.2, β =0), (α =0.2, β =0.1), … (α =1, β =1). The values that return the minimum value of RMSE are selected. To get the values of the ARIMA models' parameters (p, d, q), we first check the stationarity of the time series and determine the value of d, by performing the statistical augmented Dickey-Fuller (ADF) test [20]. The ADF test checks the null hypothesis that the time series is non-stationary and returns a probability score (p-value). A p-value of less than 5% represents that the null hypothesis can be rejected, i.e., the time series is stationary. If the p-value is greater than 5% (non-stationary time series), then the time series is differenced and the ADF test is performed again. This is repeated until the time series becomes stationary. The value of d is then equal to the number of times the series is differenced. After obtaining the value of d, we run the ARIMA (p,d,q) model using the value of d and different values of p and q from 0 and 2 at the intervals of 1 each. The values of p and q that result in the minimum RMSE are selected.

## 4.2 Experimental Results Analysis

Figure 2 shows the COVID-19 confirmed cases for the training and validation datasets for the United Arab Emirates. In addition, it indicates the number of infections forecasted using ARIMA(2,2,2) that we developed. ARIMA captures the exponentiality of the infection trend in the training dataset and forecasts accurately the infections' growth as shown in Table 1. Figure 3 shows the COVID-19 infections for the training and validation datasets for Bahrain and indicates the number of infections forecasted using ARIMA (0,2,2) that we developed. While ARIMA captures perfectly the exponentiality of the infections' growth in Bahrain during the training phase, the infections damped out during the validation phase. This leads to a higher forecasting error than in UAE (Table 1).

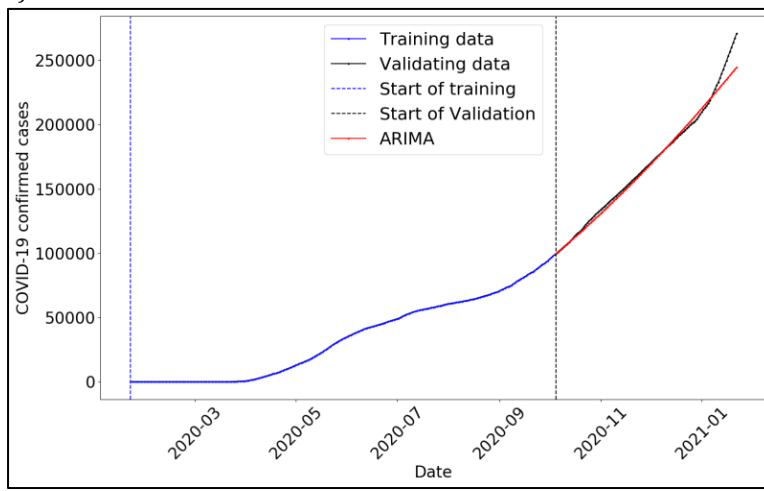

**Figure 2: Forecasting of COVID-19 infections in the United Arab Emirates using ARIMA(2,2,2)**

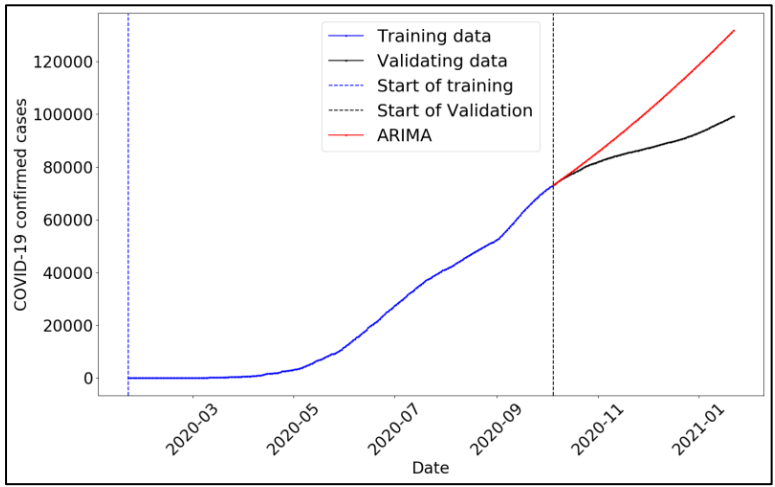

**Figure 3: Forecasting of COVID-19 infections in Bahrain using ARIMA(0,2,2)**

Figures 4, Figure 5, Figure 6 show the COVID-19 confirmed cases for the training and validation datasets for Qatar, Oman, and Kuwait respectively. They indicate the number of infections forecasted using the HLT models that we developed. The models for these countries capture the exponential + linear trend of infections' growth. Consequently, the number of infections forecasted is close to the actual ones for each country giving a higher accuracy than ARIMA for Bahrain (Table 1). This is because, HLT model fits accurately the spatial distribution of Qatar, Oman, and Kuwait data series.

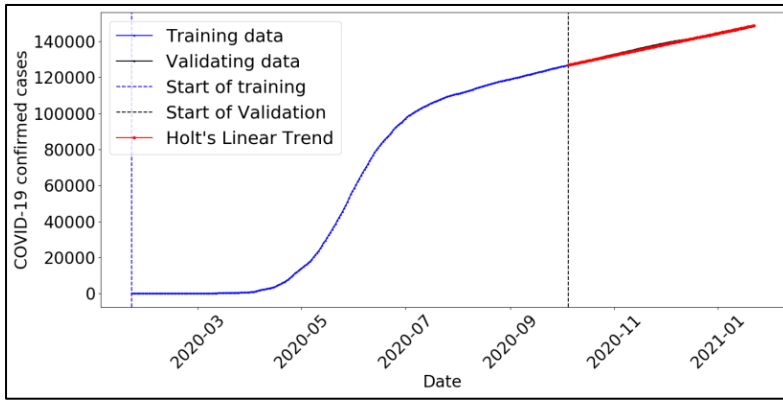

**Figure 4: Forecasting of COVID-19 infections in Qatar using Holt's Linear Trend**

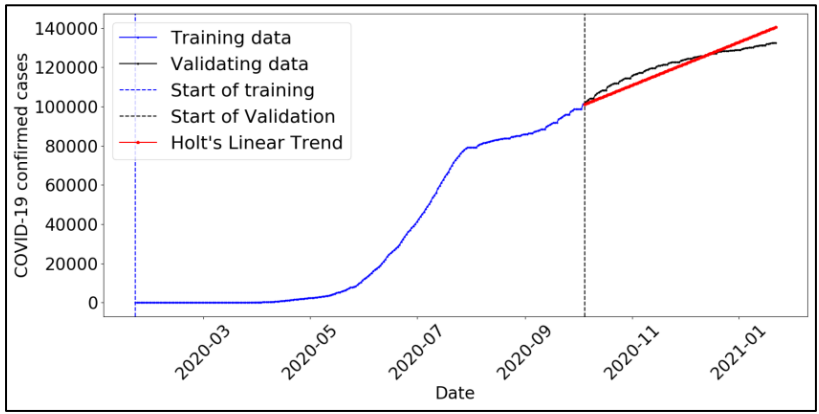

**Figure 5: Forecasting of COVID-19 infections in Oman using Holt's Linear Trend**

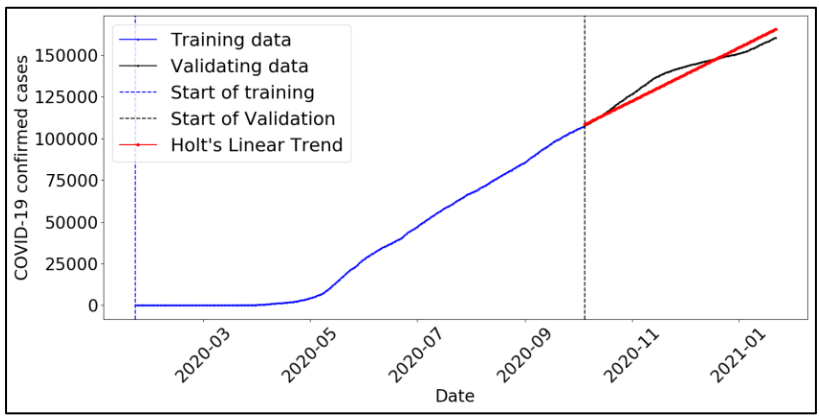

**Figure 6: Forecasting of COVID-19 infections in Kuwait using Holt's Linear Trend**

Figure 7 shows the COVID-19 confirmed cases for the training and validation datasets for Saudi Arabia, It also shows the number of infections forecasted using the DT model that we developed. The exponential + damped trend of the infections' growth in Saudi Arabia is captured by the DT model, leading to an accurate forecast as shown in Table 1.

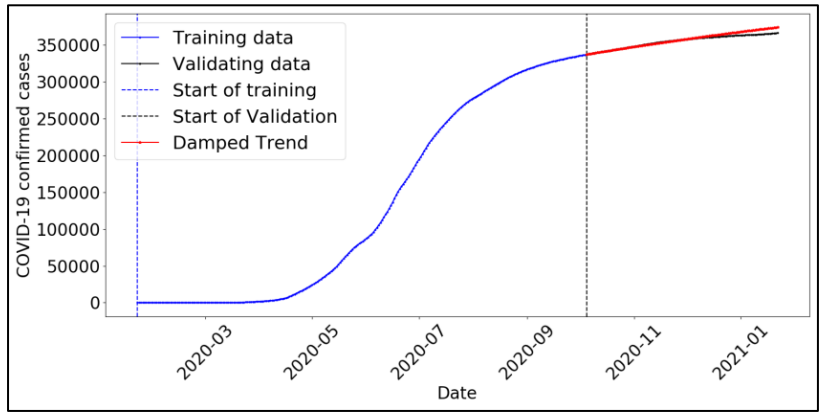

Figure 7: Forecasting of COVID-19 infections in Saudi Arabia using Damped Trend

Table 1: Root mean squared error and mean absolute percentage error for the developed models for the countries under study

| Country | Model | RMSE | MAPE |
| --- | --- | --- | --- |
| UAE | ARIMA (2,2,2) | 5905.944 | 1.6212 |
| Bahrain | ARIMA (0,2,2) | 17538.35 | 15.3610 |
| Qatar | HLT ($\alpha$=0.2, $\beta$=0.4) | 506.7492 | 0.3056 |
| Oman | HLT ($\alpha$=0.1, $\beta$=0.5) | 3946.067 | 2.8596 |
| Kuwait | HLT ($\alpha$=0.1, $\beta$=0.2) | 4150.719 | 2.5435 |
| Saudi Arabia | DT ($\alpha$=0.8, $\beta$=0.8) | 3199.276 | 0.5930 |

# 5  CONCLUSION

COVID-19 pandemic has caused more than one million deaths worldwide. Machine learning time series models have been implemented to forecast the number of COVID-19 infections in different countries. In this paper, we develop time series models to forecast infections in GCC countries. We develop a time series model for each country under study based on the spatial distribution of the infections. The developed models can be used by the government to develop effective outbreak-mitigating strategies and by the healthcare organizations to plan personnel and facilities resources efficiently.